\documentclass[10pt,twocolumn,letterpaper]{article}

\usepackage{iccv}
\usepackage{times}
\usepackage{epsfig}
\usepackage{graphicx}
\usepackage{amsmath}
\usepackage{amssymb}
\usepackage{enumitem}

\usepackage[pagebackref=true,breaklinks=true,letterpaper=true,colorlinks,bookmarks=false]{hyperref}
\usepackage[sort, numbers]{natbib}

\iccvfinalcopy 

\def\iccvPaperID{1098} 

\newcommand{\sub}{\textsubscript} 
\newcommand{\ti}{\textit} 
\newcommand{\UCF}{UCF\textunderscore CROWD\textunderscore 50}

\ificcvfinal\fi
\begin{document}

\title{Multi-Level Bottom-Top and Top-Bottom Feature Fusion for Crowd Counting}

\author{Vishwanath A. Sindagi \qquad Vishal M. Patel\\
	Department of Electrical and Computer Engineering,\\
	Johns Hopkins University, 3400 N. Charles St, Baltimore, MD 21218, USA\\
	{\tt\small \{vishwanathsindagi,vpatel36\}@jhu.edu}
}


\maketitle
\thispagestyle{empty}

\begin{abstract}

Crowd counting  presents enormous challenges  in the form of large variation in scales within images and across the dataset. These issues are further exacerbated  in highly congested scenes. Approaches  based on straightforward fusion  of multi-scale features from a deep network seem to be  obvious solutions to this problem. However,   these  fusion approaches do not yield  significant improvements in the case of crowd counting in congested scenes.  This is usually due to their limited abilities in effectively combining the multi-scale features for problems like crowd counting. To overcome this, we focus on how to efficiently  leverage information present in different layers of the network. Specifically, we present a network that involves: (i) a  multi-level  bottom-top and top-bottom fusion (MBTTBF) method to combine information from shallower to deeper layers and vice versa at multiple levels, (ii)  scale complementary feature extraction blocks (SCFB) involving cross-scale residual functions to explicitly enable flow of complementary features from  adjacent conv layers along the fusion paths.  Furthermore, in order to increase the effectiveness of the multi-scale fusion, we employ a  principled way of generating scale-aware ground-truth density maps for training.  Experiments  conducted on  three datasets that contain highly congested scenes (ShanghaiTech, \UCF, and UCF-QNRF) demonstrate that the proposed method is able to outperform  several recent methods in all the datasets. 

\end{abstract}

\section{Introduction}

Computer vision-based crowd counting \cite{li2015crowded,zhan2008crowd,idrees2013multi,zhang2015cross,zhang2016single,sindagi2017generating,sam2017switching,chan2008privacy,rodriguez2011density,zhu2014crowd,li2014anomaly,mahadevan2010anomaly} has witnessed tremendous  progress in the recent years. Algorithms developed for crowd counting have found a variety of applications such as video and traffic surveillance \cite{kang2017beyond,xiong2017spatiotemporal,onoro2016towards,zhang2017fcnrlstm,zhang2017understanding,Hsieh_2017_ICCV,toropov2015traffic}, agriculture monitoring (plant counting) \cite{lu2017tasselnet}, cell counting \cite{lempitsky2010learning}, scene understanding, urban planning and environmental survey \cite{french2015convolutional,zhan2008crowd}. 

Crowd counting from a single image, especially in congested scenes,  is a  difficult problem since it suffers from multiple issues   like  high variability in scales, occlusions,  perspective changes, background clutter, etc.  Recently, several convolutional neural network (CNN) based methods \cite{zhang2015cross,zhang2016single,sam2017switching,sindagi2017generating,liu2018leveraging,shi2018crowd_negative,shen2018adversarial,babu2018divide,cao2018scale,ranjan2018iterative}  have attempted to address  these issues with  varying degree of successes. Among these issues, the problem of scale variation has particularly received considerable attention from the research community. Scale variation typically refers to large variations in scale of the objects being counted (in this case heads)  (i) within image and (ii) across images in a dataset. 
Several other related tasks like object detection \cite{cai2016unified,li2018scale,roy2016multi,lin2017feature,hu2017finding,najibi2017ssh} and visual saliency detection \cite{hou2017deeply,ramanishka2017top,zhang2018progressive,chen2018reverse} are also affected by such effects. However, these effects are more evident especially in crowd counting in congested scenes. Furthermore, since the  annotation process for highly congested scenes is notoriously challenging,  the datasets available for  crowd counting typically provide only $x,y$ location information about the heads in the images. Since the scale labels are unavailable, training the networks to be robust to scale variations is much more challenging. In this work, we focus on addressing the issue of scale variation and missing scale information from the annotations. 

\begin{figure*}[ht!]
	\begin{center}
		\includegraphics[width=1\linewidth]{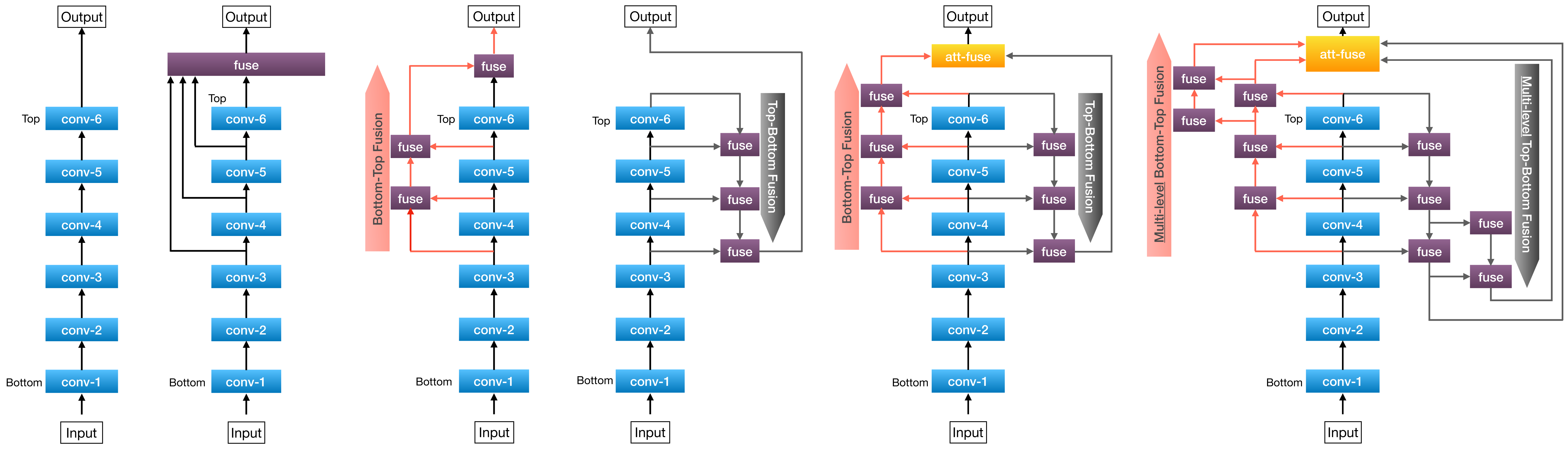}
	\end{center}
	\vskip -15pt
	\hskip40pt (a)\hskip35pt(b)\hskip60pt(c)\hskip35pt(d)\hskip80pt(e)\hskip100pt(f)
	\vskip 0pt \caption{Illustration of different multi-scale fusion architectures: (a) No fusion, (b) Fusion through concat or add, (c) Bottom-top fusion, (d) Top-bottom fusion, (e)  Bottom-top and top-bottom fusion, (f) Multi-level bottom-top and top-bottom fusion (proposed).}
	\label{fig:arch_compare}
\end{figure*}

CNNs are known to be relatively less robust  to the presence of such scale variations and hence, special techniques are required to mitigate their effects. Using features from different layers of a deep network is one approach that has been successful in  addressing this issue for other problems like object detection. It is well known that feature maps from shallower layers  encode low-level details and spatial information \cite{cai2016unified,hariharan2015hypercolumns,lin2017refinenet,ranjan2017hyperface,Yasarla_2019_CVPR}, which can be exploited to achieve better localization. However, such features are typically noisy and require further processing.  Meanwhile, deeper layers encode high-level context and semantic information  \cite{cai2016unified,hariharan2015hypercolumns,lin2017refinenet,ranjan2017hyperface} due to their larger receptive field sizes, and can aid in incorporating global context into the network. However, these features lack spatial resolution, resulting in poor localization. Motivated by these observations, we believe that high-level global semantic information and spatial localization play an important role in generating effective features for crowd counting, and hence, it is  important to fuse features  from different layers in order to achieve lower count errors. 

In order to perform an effective fusion of information from different layers of the network, we explore different fusion architectures as shown in  Fig. \ref{fig:arch_compare}\ti{(a)-(d)}, and finally arrive at our proposed method (Fig. \ref{fig:arch_compare}\ti{(f)}).  Fig. \ref{fig:arch_compare}\ti{(a)} is a typical deep network which processes the input image in a feed-forward fashion, with no explicit fusion of multi-scale features.   The network in Fig. \ref{fig:arch_compare}\ti{(b)} extracts features from multiple layers and fuses them simultaneously using a standard approach like addition or concatenation. With this configuration, the network needs to learn the importances of features from different layers automatically,  resulting in a sub-optimal fusion approach. As will be seen later in Section \ref{ssec:ablation}, this method does not produce significant improvements as compared to the base network.

To overcome this issue, one can choose to progressively incorporate detailed spatial information into the deeper layers by sequentially fusing the features from lower to higher layers (bottom-top) as shown in Fig. \ref{fig:arch_compare}\ti{(c)} \cite{sindagi2019ha}.  This fusion approach explicitly incorporates spatial context from lower layers into  the high-level features of the deeper layers.  Alternatively, a top-bottom fusion  (Fig. \ref{fig:arch_compare}\ti{(d)}) \cite{sam2018top} may be used that involves  suppressing noise in lower layers, by  propagating high-level semantic context from deeper layers into them. These approaches achieve lower counting errors as compared to the earlier configurations. However, both of these methods follow uni-directional  fusion  which may not necessarily result in optimal performance. For instance, in the case of bottom-top fusion,  noisy features also get propagated to the top layers in addition to spatial context. Similarly, in the case of top-bottom fusion, the features from the top layer may end up suppressing more than necessary details in the lower layers. Variants of these top-bottom approaches and bottom-top approaches have been proposed for other problems like semantic segmentation and object detection \cite{liu2018path,ghiasi2016laplacian,pinheiro2016learning,shrivastava2016beyond}.

Recently, a few methods \cite{yang2018multi,zhao2018defocus} have demonstrated superior performance on other tasks by using multi-directional fusion technique (Fig. \ref{fig:arch_compare}\ti{(e)}) as compared to uni-directional fusion. Motivated by the success of these methods on their respective tasks, we propose a multi-level bottom-top and top-bottom fusion (MBTTBF) technique as shown in Fig \ref{fig:arch_compare}\ti{(f)}.  By doing this, more powerful features can be learned by enabling high-level context and spatial information  to be exchanged between scales in a bidirectional manner. The bottom-top path ensures flow of spatial details into the top layer, while the top-bottom path propagates context information back into the lower layers. The feedback through both the paths ensures that minimal noise is propagated to the top layer  in the bottom-top direction, and also that the context information does not over-suppress the details in the lower layers. Hence, we are able   to effectively aggregate the advantages of different layers and suppress their disadvantages.  Note that, as compared to  existing  multi-directional fusion approaches  \cite{yang2018multi,zhao2018defocus}, we propose a more powerful fusion technique that is multi-level and  aided by scale-complementary feature extraction blocks (see Section \ref{ssec:scfb}). Additionally, the fusion process is guided by a a set of scale-aware ground-truth density maps (see Section \ref{ssec:mrf}), resulting in scale-aware features. 
 
Furthermore, we propose a scale complementary feature extraction block (SCFB) which uses cross-scale residual blocks to extract features from adjacent scales in such a way that they are complementary to each other. Traditional fusion approaches such as feature addition or concatenation are not necessarily optimal because they simple merge the features and have limited abilities to extract relevant information from different layers. In contrast, the proposed scale complementary extraction enables  the network to compute relevant features from each scale.

Lastly, we address the issue of missing scale-information in crowd-datasets by approximating the same based on the crowd-density levels and superpixel segmentation principles. Zhang \etal  \cite{zhang2016single} also estimate the scale information, however, they rely on heuristics based on the nearest number of heads. In contrast, we combine information from the annotations and super-pixel segmentation of the input image in a Markov Random Field (MRF) framework \cite{li1994markov}.   

The proposed counting method is evaluated and compared against several recent methods on  three  recent datasets that contain highly congested scenes: ShanghaiTech \cite{zhang2016single}, \UCF \cite{idrees2013multi}, and UCF-QNRF \cite{idrees2018composition}. The proposed method outperforms all existing methods by  a significant margin.  

We summarize our contributions as follows:
\begin{itemize}[topsep=0pt,noitemsep,leftmargin=*]
	\item A multi-level bottom-top and top-bottom fusion scheme to effectively merge information from multiple layers in the network. 
	\item A scale-complementary feature extraction block that is used to extract relevant features form adjacent layers of the network. 
	\item  A principled way of estimating scale-information for heads in crowd-counting datasets that involves effectively combining annotations and super-pixel segmentation in a MRF framework.  
\end{itemize}


\section{Related work}
\label{sec:related}

Compared to traditional approaches (\cite{li2008estimating,ryan2009crowd,chen2012feature,idrees2013multi,lempitsky2010learning,pham2015count,xu2016crowd}), recent methods have exploited Convolutional neural networks (CNNs) \cite{wang2015deep,zhang2015cross,sam2017switching,arteta2016counting,walach2016learning,onoro2016towards,zhang2016single,sam2017switching,sindagi2017generating,boominathan2016crowdnet} to obtain dramatic improvements in error rates. Typically, existing CNN-based methods have  focused on design of different architectures to address the issue of scale variation in crowd counting.   Switching-CNN, proposed by Babu \etal \cite{sam2017switching},   learns multiple independent regressors based on the type of image patch and has an additional switch classifier to automatically choose the appropriate  regressor for a particular input patch. More recently, Sindagi \etal \cite{sindagi2017generating} proposed Contextual Pyramid CNN (CP-CNN), where they demonstrated significant improvements by fusing local and global context through classification networks.  For a more elaborate study and discussion on these methods, interested readers are referred to  a recent survey \cite{sindagi2017survey} on CNN-based counting techniques. 

While the these methods  build  techniques that are robust to scale variations, more recent methods have focused on other aspects such as progressively increasing the capacity of the network based on dataset  \cite{babu2018divide},  use of  adversarial loss to reduce blurry effects in the predicted output maps \cite{sindagi2017generating,shen2018adversarial}, learning generalizable features via  deep negative correlation based learning \cite{shi2018crowd_negative},   leveraging unlabeled data for counting by introducing a learning to rank framework \cite{liu2018leveraging}, cascaded feature fusion \cite{ranjan2018iterative} and scale-based feature aggregation \cite{cao2018scale}, weakly-supervised learning for crowd counting \cite{sindagi2019ha}. Recently,  Idrees \etal \cite{idrees2018composition} created a new large-scale high-density crowd dataset with approximately  1.25 million head annotations and a new localization task for crowded images.

Most recently, several methods have focused on incorporating additional cues such as segmentation and semantic priors
\cite{zhao2019leveraging,wan2019residual}, attention \cite{liu2018adcrowdnet,sindagi2019ha,sindagi2019inverse}, perspective \cite{shi2019revisiting}, context information respectively \cite{liu2019context}, multiple-views \cite{zhang2019wide}  and multi-scale features  \cite{jiang2019crowd} into the network.  Wang \etal \cite{wang2019learning} introduced a new  synthetic dataset and proposed a  SSIM based CycleGAN \cite{zhu2017unpaired} to adapt the synthetic datasets to real world dataset. \\

\section{Proposed method}

In this section, we discuss details of the proposed  multi-level feature fusion scheme along with the scale complementary feature extraction blocks. This is followed by a discussion on the estimation of  head sizes using the MRF framework. 

\subsection{Multi-level bottom-top and top-bottom Fusion (MBTTBF)}

The proposed method for crowd counting is based on the recently popular density map estimation approach \cite{lempitsky2010learning,pham2015count,xu2016crowd}, where the network takes image as an input, processes it and  produces a density map. This density map indicates the per-pixel count of people in the image. The network weights are learned    by optimizing the $L_2$ error between the predicted density map and the ground truth density map. As discussed earlier, crowd counting datasets provide $x,y$ locations and these are used to create the ground-truth density maps for training by imposing 2D Gaussians  at these locations:

\begin{equation}
\label{eq:densitymap}
D_i(x) = \sum_{{x_g \in S}}\mathcal{N}(x-x_g,\sigma),
\end{equation}
where $\sigma$ is the Gaussian kernel's scale and $S$ is the list of all locations of people.   Integrating the  density map over its width and height  produces  the total count of people in the input image. 

\begin{figure}[t!]
	\begin{center}
		\includegraphics[width=1\linewidth]{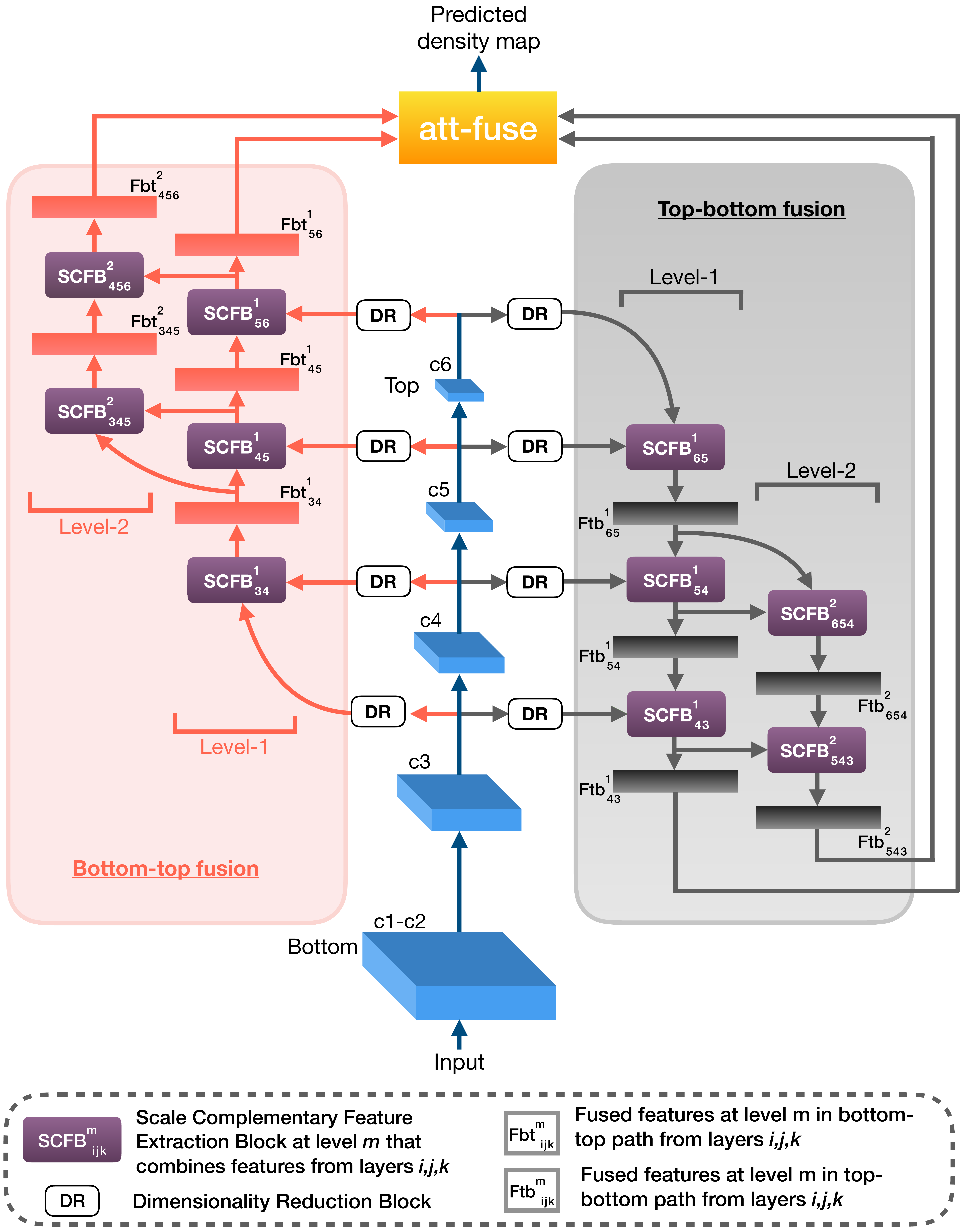}
	\end{center}
	\vskip -10pt
	\vskip 0pt \caption{Overview of the proposed multi-level top-bottom and bottom-top fusion method for crowd counting.}
	\label{fig:arch}
\end{figure}

Fig \ref{fig:arch} illustrates the overview of the proposed network. We use VGG16 \cite{simonyan2014very} as the backbone network. Conv1 - conv5 in Fig. \ref{fig:arch} are the first five convolutional layers of the VGG16 network. The last layer conv6 is defined as $\{M_2-C_{512,128,1}-R\}\footnote{ \label{fn:conv}\textit{M\sub{s}} denotes max-pooling with stride s, \textit{C\sub{N\sub{i},N\sub{o},k}} is convolutional layer (where \textit{N\sub{i}} = number of input channels, \textit{N\sub{o}} = number of output channels, \textit{k}$\times$\textit{k} = size of filter), \textit{R} is activation function (ReLU).})$. As it can be observed from this figure, the network consists of primarily three branches: (i) main branch (VGG16 backbone), (ii) multi-level bottom-top fusion branch, and (iii) multi-level top-bottom fusion branch.  The input image is passed through the main branch and multi-scale features from conv3-conv6 layers are extracted. These multi-scale features are then forwarded through  dimensionality reduction (DR) blocks that consists of 1$\times$1 conv layers to reduce the channel dimensions to 32. 


The feature maps extracted from the lower conv layers of the main branch    contain detailed spatial information which are important for accurate localization, whereas the feature maps from  higher layers contain global context and high-level information. The information contained in these different layers are fused with each other in two separate fusion branches: multi-level bottom-top branch and multi-level top-bottom branch.\\

\noindent\textbf{Multi-level bottom-top fusion:} The bottom-top branch hierarchically propagates spatial information from the bottom layers to the top layers. This branch has two levels of fusion. In the first level, features from the main branch are progressively forwarded  through a series of scale complementary feature extraction blocks ($SCFB_{34}^1$-$SCFB_{45}^1$-$SCFB_{56}^1$). First, $SCFB_{34}^1$ combines the feature maps from conv3 and conv4 to produce enriched feature maps $Fbt_{34}^1$. These features are then combined with conv5 features of the main branch through  $SCFB_{45}^1$ to produce  $Fbt_{45}^1$. Finally, these feature maps are combined with conv6 feature maps through $SCFB_{56}^1$ to produce $Fbt_{56}^1$. 

Further, we add another level of bottom-top fusion path which progressively combines features from the first level through another series of scale complementary feature extraction blocks ($SCFB_{345}^2$-$SCFB_{456}^2$).  Specifically, $Fbt_{34}^1$ and $Fbt_{45}^1$ are combined through $SCFB_{345}^2$ to produce  $Fbt_{345}^2$. Finally,  $Fbt_{345}^2$ is combined with $Fbt_{56}^1$ through $SCFB_{456}^2$ to produce $Fbt_{456}^2$. The two levels of fusion  together form a hierarchy of fusion paths. \\

\noindent\textbf{Multi-level top-bottom fusion:} The bottom-top branch while propagating spatial information to the top layers, inadvertently passes noise information as well. To overcome this, we add a top-bottom fusion path that hierarchically propagates high-level context information into the lower layers. Similar to the bottom-top path, the top-bottom path also consists of two levels of fusion.   In the first level, features from the main branch are progressively forwarded  through a series of scale complementary feature extraction blocks ($SCFB_{65}^1$-$SCFB_{54}^1$-$SCFB_{43}^1$). First, $SCFB_{65}^1$ combines the feature maps from conv6 and conv5 to produce enriched feature maps $Ftb_{65}^1$. These features are then combined with conv4 features of the main branch through  $SCFB_{54}^1$ to produce  $Ftb_{54}^1$. Finally, these feature maps are combined with conv3 feature maps through $SCFB_{43}^1$ to produce $Ftb_{43}^1$. 

The second level of bottom-top fusion path  progressively combines features from the first level through another series of scale complementary feature extraction blocks ($SCFB_{654}^2$-$SCFB_{543}^2$). Specifically, $Ftb_{65}^1$ and $Ftb_{54}^1$ are combined through $SCFB_{654}^2$ to produce  $Ftb_{654}^2$. Finally,  $Ftb_{654}^2$ is combined with $Ftb_{43}^1$ through $SCFB_{543}^2$ to produce $Fbt_{543}^2$. Again, the two levels of fusion  together form a hierarchy of fusion paths in the top-bottom module. \\

\noindent\textbf{Self attention-based fusion:} The features produced by the bottom-top fusion ($Fbt_{56}^1$ and $Fbt_{456}^2$), although refined, may contain some unnecessary background clutter. Similarly, the features ($Ftb_{43}^1$ and $Ftb_{543}^2$) produced by the top-bottom fusion may over suppress the detail information in the lower layers. In order to further suppress the background noise in the bottom-top path and avoid over-suppression of detail information due to the top-bottom path, we introduce a self-attention based fusion module at the end that combines feature maps from the two fusion paths. Given the set of feature maps ($Fbt_{56}^1 $ , $Fbt_{456}^2$, $Ftb_{43}^1$ and $Ftb_{543}^2$) from the fusion branches, the attention module concatenates them and forwards them through a set of conv layers ($\{C_{128,16,3}-R-\{C_{16,4,1}\}^\textsc{\ref{fn:conv}}$) and a sigmoid  layer to  produces an  attention maps with four channels, with each channel specifying the importance of the corresponding feature map from the fusion branch.  The attention maps are calculated as follows: $A = sigmoid(cat(F_{56}^1  , F_{456}^2, F_{43}^1,F_{543}^2))$.

\begin{figure}[ht!]
	\begin{center}
		\includegraphics[width=0.95\linewidth]{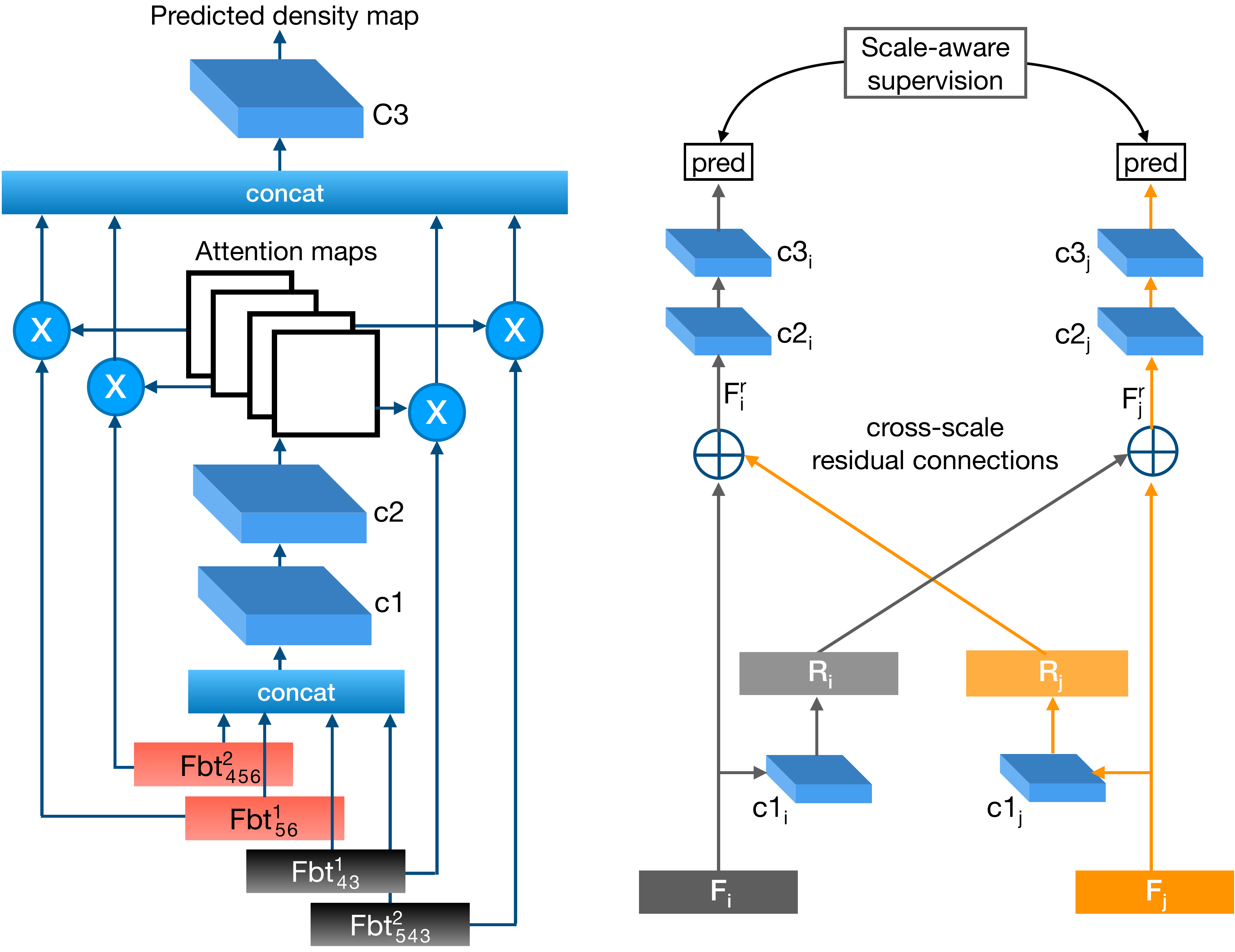}\\
		(a)\hskip100pt(b)
		 
	\end{center}
	\vskip -12pt
	\vskip 0pt \caption{(a)Attention fuse module. (b) Scale complementary feature extraction block (SCFB).}
	\label{fig:att_fuse}
\end{figure}

These attention maps are then multiplied element-wise to produce the final feature map: $F_f = A^1\odot F_{56}^1 + A^2\odot F_{456}^2 + A^3\odot F_{43}^1 + A^4\odot F_{543}^2$, where $\odot$ denotes element-wise multiplication. This self-attention module effectively combines the advantages of the two paths, resulting in more powerful and enriched features. Fig. \ref{fig:att_fuse}\ti{(a)} shows the self-attention block used to combine different feature maps. The final features $F_f$ are then forwarded through 1$\times$1 conv layer to produce the density map $Y_{pred}$.

\begin{figure}[t!]
	\begin{center}
		\includegraphics[width=1\linewidth]{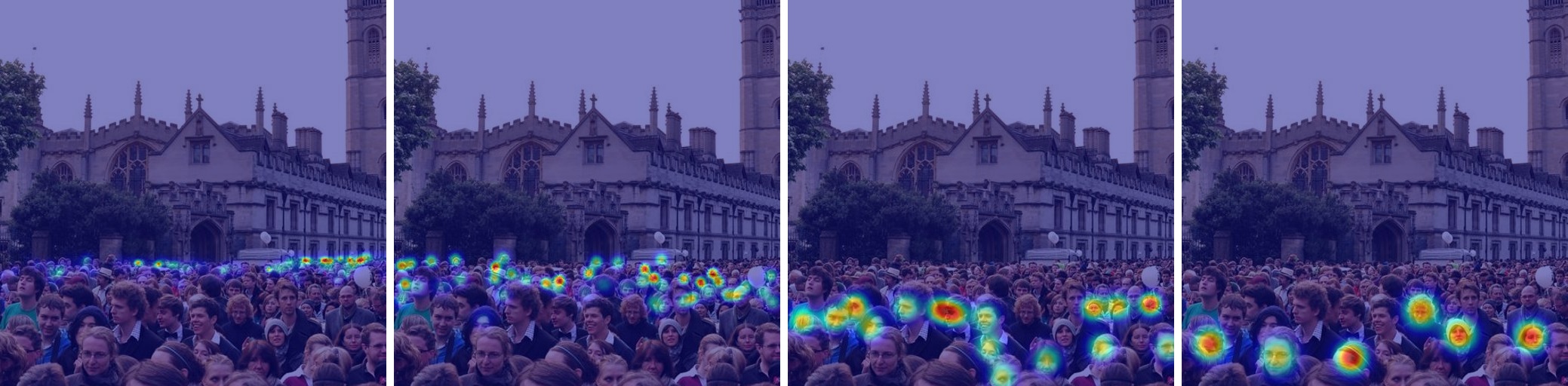}\\
		\includegraphics[width=1\linewidth]{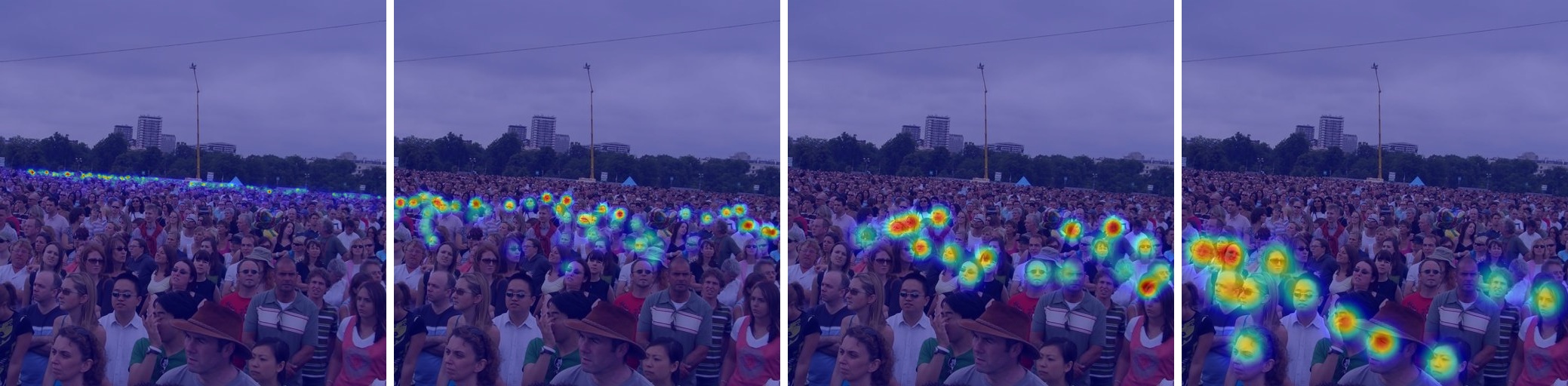}
		
	\end{center}
	\vskip -10pt
	\vskip 0pt \caption{Scale aware ground truth density maps imposed on the input image. The overall density map is divided into four maps based on the size/scale of the heads. The first image (leftmost) has density corresponding to the smallest set of heads, whereas the last image (rightmost) has densities corresponding to the largest set of heads.}
	\label{fig:scale_aware_density}
\end{figure}

\subsection{Scale complementary feature extraction block (SCFB)}
\label{ssec:scfb}

In this section, we describe the scale complementary feature extraction block that is used to combine features from adjacent layers in the network. Existing methods such as feature addition or concatenation are limited in their abilities to learn complementary features. This is because   features of adjacent layers are correlated, and this results in some ambiguity in the fused features. To address this issue, we introduce scale complementary feature extraction block as shown in Fig. \ref{fig:att_fuse}\ti{(b)}. This block  enables extraction of complementary features from each of the scales being fused.   The initial conv layers $c1_i, c1_j, c2_i, c2_j$ in Fig. \ref{fig:att_fuse}\ti{(b)} are defined as $\{C_{32,32,3}-R\}^\textsc{\ref{fn:conv}}$, where as the final conv layers  $c3_i, c3_j$ are defined as $\{C_{32,1,1}-R\}^\textsc{\ref{fn:conv}}$.

The SCFB consists of cross-scale residual connections ($R_i$ and $R_j$) which are followed by a set of conv layers. The individual branches in the SCFB are supervised by scale-aware supervision (which is now possible due to the scale estimation framework discussed in Section \ref{ssec:mrf}). More specifically, in order to combine feature maps $F_i, F_j$ from layers $i, j$, first the corresponding cross-scale residual features $F_i^r, F_j^r$ are estimated and added to the original feature maps  $F_i, F_j$ to produce  $\hat{F}_i, \hat{F}_j$, \textit{i.e.,} $\hat{F}_i = F_i + F_j^r$ and $\hat{F}_j = F_j + F_i^r$. These features are then forwarded through a set of conv layers, before being supervised by the scale-aware ground-truth density maps $Y_i^s, Y_j^s$. By adding these intermediate supervisions and introducing the cross-scale residual connections, we are able to compute complementary features from the two scales in the form of residuals. This reduces the ambiguity as compared to the existing fusion methods. For example, if a feature map $F_i$ from a particular layer/scale $i$ is sufficient enough to obtain perfect prediction, then the residual $F_j^r$ is simply driven towards zero. Hence,  involving residual functions reduces the ambiguity as compared to the existing fusion techniques. 

In order to supervise the SCFBs, we create scale-aware ground-truth density maps  based on the scales/sizes estimated as described in Section \ref{ssec:mrf}. Annotations  in a particular image are divided into four categories based on the corresponding head sizes, and these four categories are used to create four separate ground-truth density maps ($Y_3^s, Y_4^s, Y_5^s and Y_6^s$) for a particular image. Fig. \ref{fig:scale_aware_density} shows the  four scale-aware ground-truth density maps for two sample images. It can be observed that the first ground-truth (left) has labels corresponding to the smallest heads, where as the  last ground-truth (right) has labels corresponding to the largest heads.  These maps  ($Y_3^s, Y_4^s, Y_5^s and Y_6^s$) are used to provide intermediate supervision to feature maps coming from conv layers 3,4,5 and 6 coming from the main branch in SCFBs. 

\subsection{Head size estimation using MRF framework}
\label{ssec:mrf}

\begin{figure}[ht!]
	\begin{center}
		\includegraphics[width=1\linewidth]{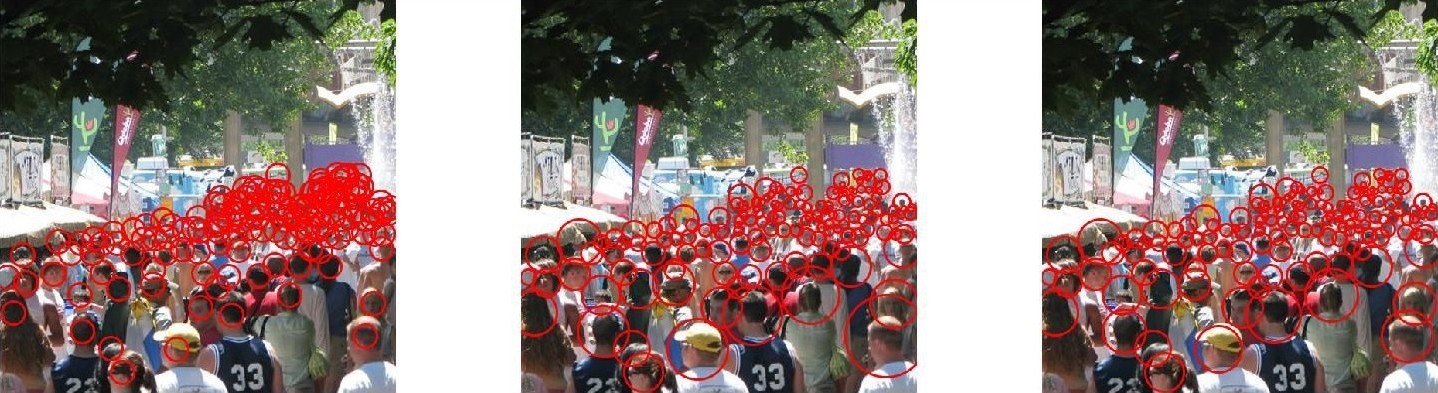}\\
		\includegraphics[width=1\linewidth]{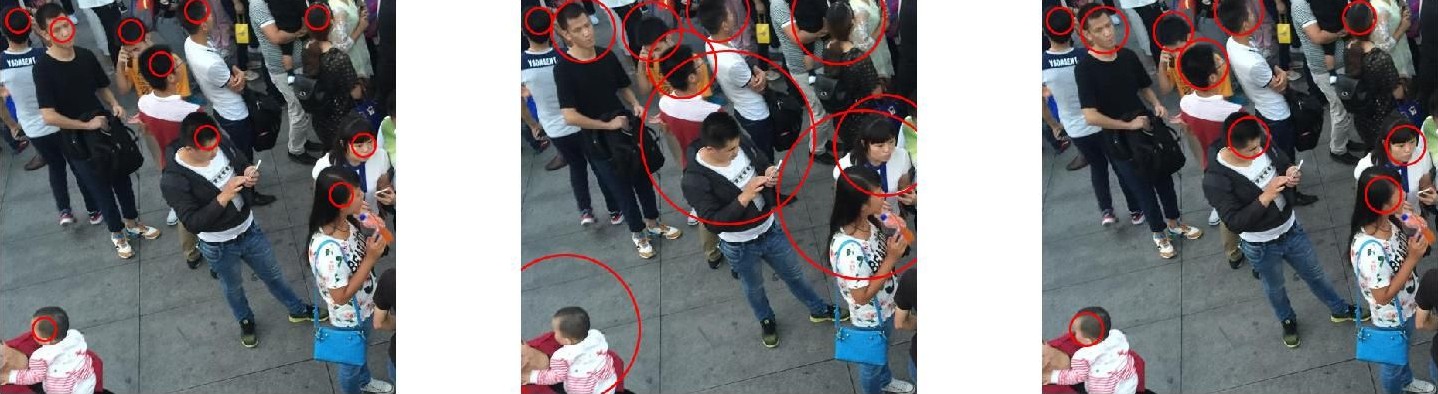}\\
		(a)\hskip75pt(b)\hskip75pt(c)
	\end{center}
	\vskip -10pt
	\vskip 0pt \caption{Scale estimation comparison. Scale estimated using (a) Constant scale (b) Nearest neighbours (c) Our method.}
	\label{fig:scale}
\end{figure}

As discussed earlier, the ground truth density maps for training the CNNs are created by imposing 2D Gaussians  at the head locations (Eq. \eqref{eq:densitymap}) provided in the dataset. The scale/variance of these Gaussians needs to be decided based on the heads size. Existing methods either assume constant variance \cite{sindagi2017generating} or estimate  the variance based on the number of nearest heads \cite{zhang2016single}.  Assuming constant variance results in ambiguity in the density maps and hence, prohibits the network to learn scale relevant features. Fig. \ref{fig:scale}\ti{(a)} shows the scales for annotations assuming constant variance. On the other hand, estimating the variance based on nearest neighbours leads to better results in regions of high density. However, in regions of low density, the estimates are incorrect leading to ambiguity in such regions (as shown in Fig. \ref{fig:scale}\ti{(b)}). 

To overcome these issues, we propose a principled way of estimating the scale or variance by considering the input images   which were not exploited earlier. We leverage color cues from the input image and combine them with the annotation data to better estimate the scale. Specifically, we first over-segment the input image using a super-pixel algorithm (SLIC \cite{achanta2010slic}) and then combine with watershed segmentation \cite{beucher1992watershed} resulting from the distance transform of the head locations in an MRF framework.  The size of the segments resulting from this procedure are then used to estimate the scale of the corresponding head lying in that segment. Fig. \ref{fig:scale}\ti{(c)}  shows the scales/variances estimated using the proposed method. It can be observed that this method performs better in both sparse and dense regions. 

\section{Details of implmentation and training}

\noindent The network weights are optimized in and end-to-end fashion. We use  Adam optimizer with a learning rate of 0.00005 and a momentum of 0.9. 
We add random noise and perform random flipping of images for data augmentation. We use mean absolute error ($MAE$) and mean squared error ($MSE$) for evaluating the network performance. These metrics are defined as: $MAE = \frac{1}{N}\sum_{i=1}^{N}|y_i-y'_i|$ and  $MSE = \sqrt{\frac{1}{N}\sum_{i=1}^{N}|y_i-y'_i|^2}$ respectively, where  $N$ is the total number of test images, $y_i$ is the ground-truth/target count of people in the image and $y'_i$ is the predicted count of people  in to the $i^{th}$ image. Supervision is provided to the network at the final level as well as at intermediate levels  in the SCFBs using Euclidean loss. At the final level, the network is supervised by the overall density map (consisting of annotations corresponding to all the heads), whereas the paths in the SCFBs are supervised by the corresponding scale-aware ground-truths. 

 

\section{Experiments and results}
In this section, we first analyze the different components involved in the proposed network through an  ablation study.  This is followed by a detailed evaluation of the proposed method and  comparison with several recent state-of-the-art methods.

\subsection{Datasets}
\label{ssec:datasets}

We use three different congested crowd scene datasets (ShanghaiTech \cite{zhang2016single}, \UCF \cite{idrees2013multi} and UCF-QNRF \cite{idrees2018composition}) for evaluating the proposed method. The ShanghaiTech  \cite{zhang2016single} dataset  contains 1198 annotated images with a total of 330,165 people. This dataset consists of two parts: Part A with 482 images and Part B with 716 images. Both parts are further divided into training and test datasets with training set of Part A containing 300 images and that of Part B containing 400 images. The UCF\textunderscore CC\textunderscore 50 is an extremely challenging dataset introduced by Idrees \etal \cite{idrees2013multi}. The dataset contains 50 annotated images of different resolutions and aspect ratios crawled from the internet. The UCF-QNRF \cite{idrees2018composition} dataset, introduced recently by Idrees \etal, is a large-scale crowd dataset containing 1,535 images with 1.25 million annotations. The images are of high resolution and are collected under a diverse backgrounds such as buildings, vegetation, sky and roads. The training and test sets in this dataset consist of 1201 and 334 images, respectively. 


\subsection{Ablation Study}
\label{ssec:ablation}

We perform a detailed ablation study to understand the effectiveness of  various fusion approaches described earlier. The ShanghaiTech Part A and UCF-QNRF datasets contain  different conditions 	 such as high variability in scale, occluded objects and  large  crowds, \etc.  Hence, we used these datasets for conducting the ablations. The following configurations were trained and evaluated:

\begin{figure*}[ht!]
	\begin{center}
		\includegraphics[width=.99\linewidth]{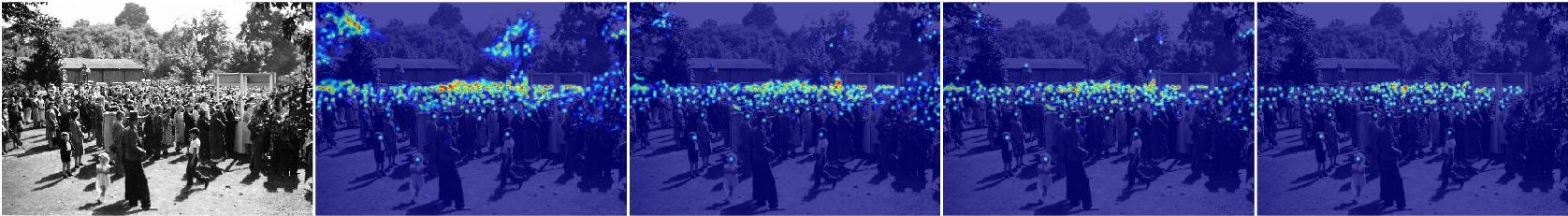}
		\includegraphics[width=.99\linewidth]{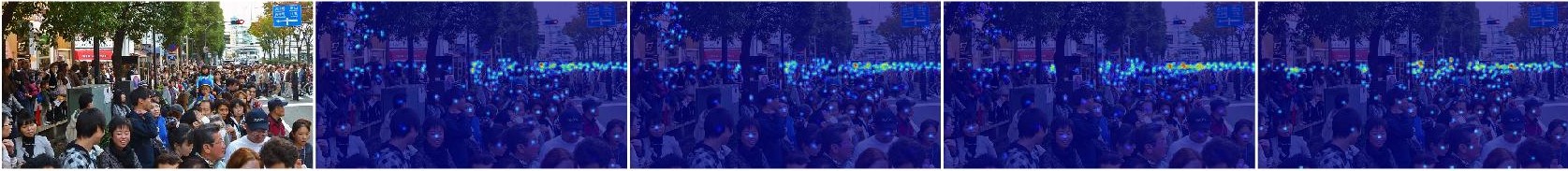}
		\includegraphics[width=.999\linewidth]{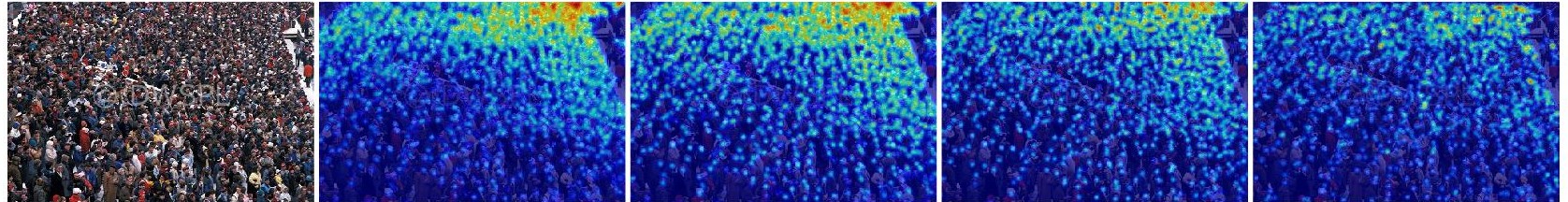}\\
		(a)\hskip85pt(b)\hskip85pt(c)\hskip85pt(d)\hskip85pt(e) 
		
	\end{center}
	\vskip -10pt
	\vskip 0pt \caption{Ablation study results: (a) Input, (b) Simple feature concatenation (experiment-ii), (c) Bottom-top and top-bottom fusion (experiment - vi), (d) MBTTF (experiment - viii), (e) Ground-truth density map. }
	\label{fig:ablation}
		\vskip -10pt
\end{figure*}

\noindent(i) \textit{Baseline}: VGG16 network with   $conv6$ at the end (Fig. \ref{fig:arch_compare}(a)), \\
\noindent(ii) \textit{Baseline + fuse-a}: Baseline network with multi-scale feature fusion using feature addition  (Fig. \ref{fig:arch_compare}(b)), \\
\noindent(iii) \textit{Baseline + fuse-c}: Baseline network with multi-scale feature fusion using feature concatenation  (Fig. \ref{fig:arch_compare}(b)),\\ 
\noindent(iv) \textit{Baseline + BT + fuse-c}: Baseline network with bottom-top multi-scale feature fusion using feature concatenation  (Fig. \ref{fig:arch_compare}(c)), \\
\noindent(v) \textit{Baseline + TB + fuse-c}: Baseline network with top-bottom multi-scale feature fusion using feature concatenation  (Fig. \ref{fig:arch_compare}(d)), \\
\noindent(vi) \textit{Baseline +  BTTB + fuse-c}: Baseline network with bottom-top and top-bottom multi-scale feature fusion using feature concatenation  (Fig. \ref{fig:arch_compare}(e)), \\
\noindent(vii) \textit{Baseline + MBTTB + fuse-c}: Baseline network with multi-level bottom-top and top-bottom multi-scale feature fusion using feature concatenation  (Fig. \ref{fig:arch_compare}(f)), \\
\noindent(viii) \textit{Baseline + MBTTB + SCFB-NS}: Baseline network with multi-level bottom-top and top-bottom multi-scale feature fusion using SCFB, without using scale-aware supervision (Fig. \ref{fig:arch}) \\
\noindent(ix) \textit{Baseline + MBTTB + SCFB}: Baseline network with multi-level bottom-top and top-bottom multi-scale feature fusion using SCFB (Fig. \ref{fig:arch}) \\

\begin{table}[t!]
	\caption{Ablation study results.}
	\label{tab:ablation}
	\resizebox{1\linewidth}{!}{
		\begin{tabular}{|l|c|c|c|c|}
			\hline
			Dataset & \multicolumn{2}{c|}{Shanghaitech-A\cite{zhang2016single}} & \multicolumn{2}{c|}{UCF-QNRF\cite{idrees2018composition}} \\ \hline
			Method & MAE & MSE & MAE & MSE \\ \hline
			Baseline (Fig. \ref{fig:arch_compare}a) & 78.3  & 126.6 & 150.2  & 220.1 \\   
			Baseline + fuse-a (Fig. \ref{fig:arch_compare}b) &  73.6 & 118.4 & 140.3  & 210.8  \\  
			Baseline + fuse-c (Fig. \ref{fig:arch_compare}b) & 73.4 &  115.6 &  135.2 &  200.2 \\ 
			Baseline + BT + fuse-c (Fig. \ref{fig:arch_compare}c) &  68.1 & 122.2 &  114.1 &  185.2\\  
			Baseline + TB + fuse-c (Fig. \ref{fig:arch_compare}d) & 70.2  & 118.5 &120.1  &  188.1\\  
			Baseline + BTTB + fuse-c (Fig. \ref{fig:arch_compare}e) &66.9  & 112.2 & 115.4 & 174.5 \\ 
			Baseline + MBTTB + fuse-c (Fig. \ref{fig:arch_compare}f) & 63.2  & 108.5 & 105.5 & 169.5  \\  
			Baseline + MBTTB + SCFB-NS (Fig. \ref{fig:arch}) &  62.5 & 105.1 & 102.1 & 168.1 \\ 
			Baseline + MBTTB + SCFB (Fig. \ref{fig:arch}) &  60.2 & 94.1 & 97.5 & 165.2 \\ \hline
		\end{tabular}
	}
\end{table}

The quantitative results of the ablation study are shown in Table \ref{tab:ablation}. As it can be observed, simple fusion scheme of   addition/concatenation (experiments (i) and (ii)) of multi-scale features at the end, does not yield significant improvements as compared to the baseline network.  This is due to the reason that in case of feature fusion at the end, the supervision directly affects the initial conv layers in the main branch, which may not be necessarily optimal. 

However, when the features are fused in either bottom-top/top-bottom fashion, the results improve considerably, when compared to the baseline. Since this kind of fusion sequentially propagates the information in a particular direction, the initial conv layers do not get affected directly.  The bottom-top and top-bottom (experiment (vi))  further improves the  performance. The  multi-level bottom-top and top-bottom configuration, in which an additional level of bottom-top and top-bottom fusion path is added  (experiment-vii), reduces the count error further, signifying the importance of the multi-level fusion paths. 


Next, we replace the fusion blocks in experiment-vii with the SCFB blocks, which amounts to the proposed method as shown in Fig. \ref{fig:arch} (experiment viii). However, the SCFB blocks are not supervised by the scale-aware ground-truths. The use of these blocks enables the network to propagate relevant and complementary features along the fusion paths, thus leading to improved performance. Finally, we provide scale-aware ground-truth as supervision signal to the SCFB blocks (experiment - ix), which results in further improvements as compared to without scale-aware supervision. 

Fig. \ref{fig:ablation} shows qualitative results for different fusion configurations. Due to space constraints and also to explain better, we show the results of experiments (iii) \textit{Baseline + fuse-c}, (vi) \textit{Baseline +  BTTB + fuse-c}, (ix) \textit{Baseline + MBTTB + SCFB} only.   It can be observed from Fig. \ref{fig:ablation}\ti{(b)}, that simple concatenation of feature  maps  results in lot of background noise and loss of details in the final predicted density map, indicating that such an approach is not effective. The bottom-top and top-bottom approach, shown in Fig. \ref{fig:ablation}\ti{(c)} results in the refined density maps, however, they still contain some amount of noise and loss of details. Lastly, the results of experiment (ix) as shown in Fig. \ref{fig:ablation}\ti{(d)} which have more details where necessary with much lesser background clutter as compared to earlier configurations.

\subsection{Comparison with recent methods}

In this section, we present the  results of the proposed method and compare them  with several recent  approaches on the three different datasets described in Section \ref{ssec:datasets}. 

Comparison of results the ShanghaiTech and \UCF\; datasets are presented in Table \ref{tab:shtech} and \ref{tab:ucf_cc} respectively. The proposed method achieves the best results among  all the existing methods on the ShanghaiTech Part A dataset and the \UCF\;  dataset. On the ShanghaiTech B dataset and \UCF dataset, our method achieves a close $2^{nd} $ position, only behind CAN \cite{liu2019context}.

\begin{table}[h!]
	\centering
	\caption{Comparison of results on ShanghaiTech  \cite{zhang2016single}.}
	\label{tab:shtech}
	\resizebox{1\linewidth}{!}{
		\begin{tabular}{|l|c|c|c|c|}
			\hline
			& \multicolumn{2}{c|}{Part A} & \multicolumn{2}{c|}{Part B} \\ \hline
			Method          & MAE          & MSE          & MAE          & MSE          \\ \hline
			Switching-CNN \cite{sam2017switching}    (CVPR-17)       & 90.4        & 135.0        & 21.6         & 33.4         \\ 
			TDF-CNN \cite{sam2018top}  (AAAI-18)         & 97.5       & 145.1        & 20.7         & 32.8         \\ 
			CP-CNN \cite{sindagi2017generating} (ICCV-17)           & 73.6        & 106.4        & 20.1         & 30.1         \\ 
			IG-CNN \cite{babu2018divide}   (CVPR-18)        & 72.5        & 118.2        & 13.6         & 21.1          \\ 
			Liu \etal \cite{liu2018leveraging}  (CVPR-18)         & 73.6        & 112.0        & 13.7         & 21.4        \\
			CSRNet \cite{li2018csrnet} (CVPR-18)          & 68.2        & 115.0        & 10.6         & 16.0    \\ 
			SA-Net  \cite{cao2018scale} (ECCV-18)           & {67.0}        & {104.5}        &  {{8.4}}         &  {{13.6}}       \\
			ic-CNN \cite{ranjan2018iterative}  (ECCV-18)         & 69.8        & 117.3        & 10.7         & 16.0       \\
			ADCrowdNet \cite{liu2018adcrowdnet} (CVPR-19)           & {63.2}        & {98.9}        & {{8.2}}         &  15.7       \\
			RReg \cite{wan2019residual} (CVPR-19)      & {63.1}        & {96.2}        & {{8.7}}         &  {{13.5}}       \\
			CAN  \cite{liu2019context} (CVPR-19) & {\textbf{61.3}}        & {{100.0}}        & \underline{\bf{7.8}}         & \underline{\bf{12.2}}\\
			Jian \etal  \cite{jiang2019crowd} (CVPR-19) & {{64.2}}        & {{109.1}}        & {{8.2}}         & {\bf{12.8}}\\
			HA-CCN  \cite{sindagi2019ha} (TIP-19) & {{62.9}}        & {\textbf{94.9}}        & {{8.1}}         & {{13.4}}\\
			MBTTBF-SCFB (proposed) & \underline{\textbf{60.2}}        & \underline{\textbf{94.1}}        & \textbf{8.0}         & {15.5}         \\ \hline
		\end{tabular}

	}

\end{table}

Results on the recently released large-scale UCF-QNRF \cite{idrees2018composition} dataset are shown in Table \ref{tab:resultsucf}. We compare our results with several recent   approaches. The proposed achieves the best results as compared to other recent methods on this complex dataset, thus demonstrating the  significance of the proposed multi-level fusion method.  

\begin{figure}[t]	
	\centering	
	\includegraphics[width=0.323\linewidth]{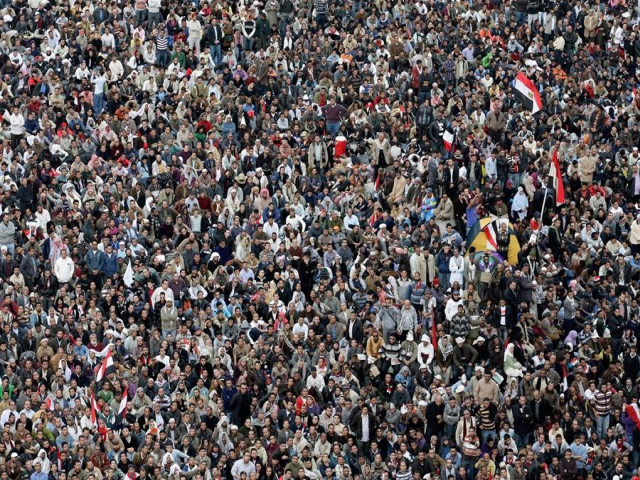}
	\includegraphics[width=0.323\linewidth]{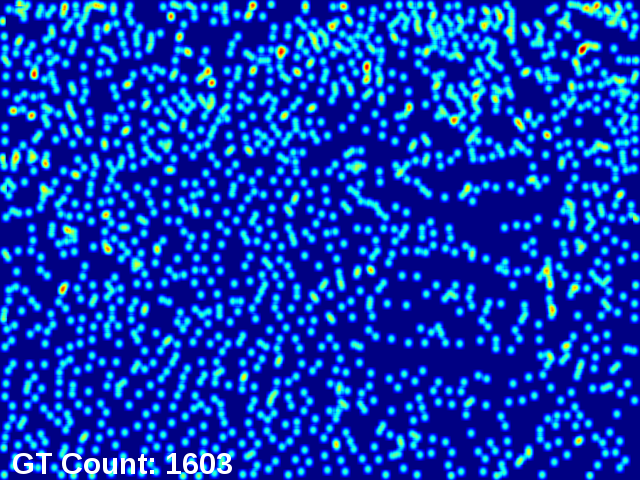}
	\includegraphics[width=0.323\linewidth]{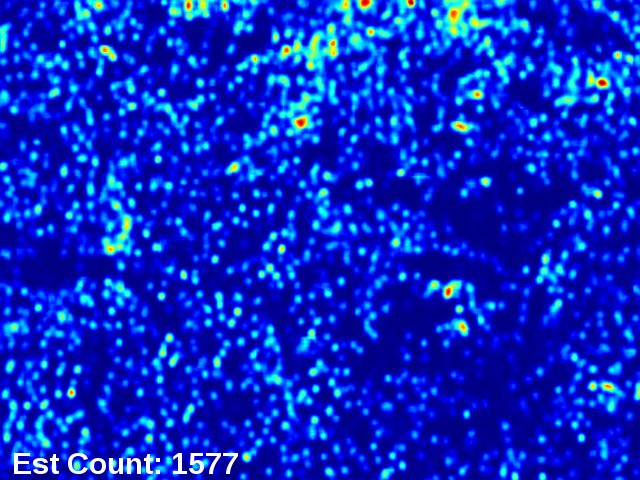}
	
	\includegraphics[width=0.323\linewidth]{figures/sa/IMG_40.jpg}
	\includegraphics[width=0.323\linewidth]{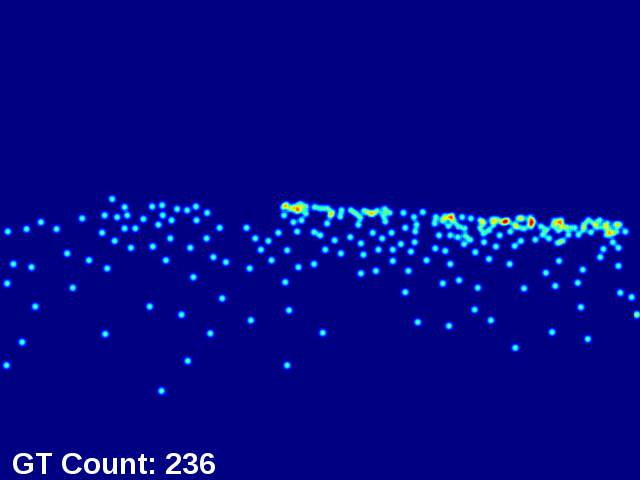}
	\includegraphics[width=0.323\linewidth]{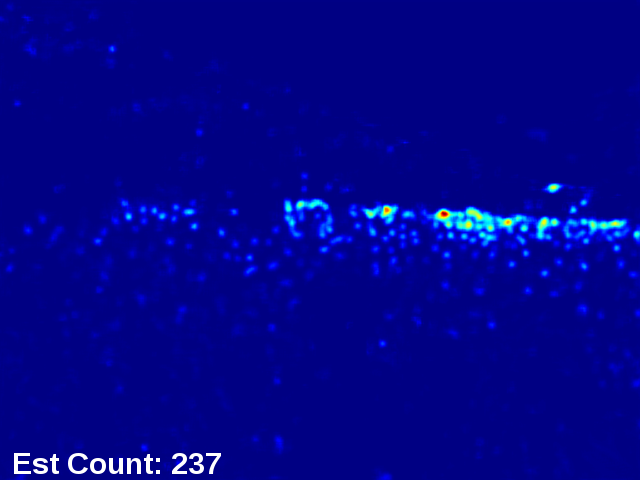}	
	\\
	
	\vskip -3pt\caption{Qualitative results of the proposed method on  ShanghaiTech \cite{zhang2016single}  \textit{First column:} Input. \textit{Second column:} Ground truth \textit{Third column:} Predicted density map.}
	\label{fig:shtechresults1}	
\end{figure}

Qualitative results for sample images from the ShanghaiTech dataset are presented in Fig. \ref{fig:shtechresults1}.

\begin{table}[h!] 
	\centering 		
	\caption{Comparison of results on   \UCF \cite{idrees2015detecting}.}
	\label{tab:ucf_cc}
	\resizebox{0.9\linewidth}{!}{
		\begin{tabular}{|l|c|c|}
			\hline
			&  \multicolumn{2}{|c|}{\UCF}\\ \hline
			Method               & MAE          & MSE         \\ \hline
			Switching-CNN \cite{sam2017switching}  (CVPR-17)                 & 318.1         & 439.2        \\ 
		    TDF-CNN \cite{sam2018top}  (AAAI-18)                & 354.7         & 491.4        \\ 
			CP-CNN \cite{sindagi2017generating}  (ICCV-17)               & 295.8         & {{320.9 }}       \\ 
			IG-CNN \cite{babu2018divide}    (CVPR-18)             & 291.4         & 349.4        \\ 
			D-ConvNet \cite{shi2018crowd_negative}  (CVPR-18)        & 288.4         & 404.7         \\ 
			Liu \etal \cite{liu2018leveraging}    (CVPR-18)         & 289.6         & 408.0       \\
			CSRNet \cite{li2018csrnet}  (CVPR-18)        & 266.1         & 397.5       \\
			ic-CNN \cite{ranjan2018iterative}    (ECCV-18)         & 260.9         & 365.5       \\
			SA-Net-patch \cite{cao2018scale}   (ECCV-18)          &  258.5       & 334.9  \\
			ADCrowdNet \cite{liu2018adcrowdnet} (CVPR-19)       & {266.4}        & 358.0   \\
			CAN \cite{liu2019context}  (CVPR-19)  &    \underline{\bf{212.2}}         & \underline{\bf{243.7}}       \\ 
			Jian \etal \cite{jiang2019crowd}  (CVPR-19)  &    {{249.9}}         & {{354.5}}       \\
			HA-CCN \cite{sindagi2019ha}  (TIP-19)  &    {{256.2}}         & 348.4       \\ 
			MBTTBF-SCFB (ours) &  {\bf{233.1}}         & \bf{300.9}       \\ \hline
		\end{tabular}
	}
	
\end{table} 

\begin{table}[ht!]
	\centering
	\caption{Comparison of results on the UCF-QNRF datastet \cite{idrees2018composition}. }
	\label{tab:resultsucf}
	\resizebox{0.9\linewidth}{!}{
		\begin{tabular}{|l|c|c|}
			\hline
			Method & MAE & MSE \\			\hline
			CMTL  \cite{sindagi2017cnnbased} (AVSS-17) & 252.0 & 514.0 \\
			MCNN \cite{zhang2016single} (CVPR-16)  & 277.0 & 426.0\\
			Switching-CNN \cite{sam2017switching}  (CVPR-17) & 228.0 & 445.0 \\
			Idrees \etal \cite{idrees2018composition}  (ECCV-18) & {132.0} & {{191.0}} \\
			Jian \etal \cite{jiang2019crowd} (CVPR-19) & {{113.0}} & {{188.0 }}\\
			CAN \cite{liu2019context} (CVPR-19) & {{107.0}} & {{183.0 }}\\
			HA-CCN \cite{sindagi2019ha} (TIP-19) & {{118.1}} & {{180.4 }}\\
			MBTTBF-SCFB (ours) & {\textbf{97.5}} & {\bf{165.2 }}\\
			\hline
		\end{tabular}
	}
\end{table}

%
%
%

\section{Conclusion}
We presented a multi-level bottom-top and top-bottom  fusion scheme for overcoming the issues of scale variation that adversely affects crowd counting in congested scenes. The proposed method  first extracts a set of scale-complementary features from adjacent layers before propagating them hierarchically in bottom-top and top-bottom fashion. This results in a more effective fusion of  features from multiple layers of the backbone network.  The effectiveness of the proposed fusion scheme is further enhanced by using ground-truth density maps that are created in a principled way by combining information from the image and location annotations in the dataset. In comparison to existing fusion schemes and state-of-the-art counting methods, the proposed approach is able to achieve significant improvements when evaluated on three popular crowd counting datasets. 

\section*{Acknowledgment}
\noindent This work was supported by the NSF grant 1922840.

{\small
\bibliographystyle{ieee_fullname}
\bibliography{egbib}
}

\end{document}